\documentclass[acmsmall]{acmart}

\setcopyright{none}
\copyrightyear{2024}
\acmDOI{XXXXXXX.XXXXXXX}

\usepackage{tabularx}
\usepackage{wrapfig,lipsum,booktabs}
\usepackage{xcolor,colortbl}

\begin{document}

\title{Transfer Learning Applied to Computer Vision Problems: Survey on Current Progress, Limitations, and Opportunities}

\author{Aaryan Panda}
\email{aaryanpanda001@gmail.com}
\affiliation{
 \institution{Hillsborough High School (IB)}
  \streetaddress{5000 N Central Ave}
  \city{Tampa}
  \state{Florida}
  \country{USA}
  \postcode{33603}
}

\author{Damodar Panigrahi}
\email{dp1657@msstate.edu}
\affiliation{
 \institution{Mississippi State University}
  \streetaddress{P.O. Box 1212}
  \city{Mississippi State}
  \state{Mississippi}
  \country{USA}
  \postcode{39762}
}

\author{Shaswata Mitra}
\email{sm3843@msstate.edu}
\affiliation{
 \institution{Mississippi State University}
  \streetaddress{P.O. Box 1212}
  \city{Mississippi State}
  \state{Mississippi}
  \country{USA}
  \postcode{39762}
}

\author{Sudip Mittal}
\email{mittal@cse.msstate.edu}
\affiliation{%
 \institution{Mississippi State University}
  \streetaddress{P.O. Box 1212}
  \city{Mississippi State}
  \state{Mississippi}
  \country{USA}
  \postcode{39762}
}

\author{Shahram Rahimi}
\email{rahimi@cse.msstate.edu}
\affiliation{%
 \institution{Mississippi State University}
  \streetaddress{P.O. Box 1212}
  \city{Mississippi State}
  \state{Mississippi}
  \country{USA}
  \postcode{39762}
}

\renewcommand{\shortauthors}{Panda et al.}



\begin{abstract}
The field of Computer Vision (CV) has faced challenges. Initially, it relied on handcrafted features and rule-based algorithms, resulting in limited accuracy. The introduction of machine learning (ML) has brought progress, particularly Transfer Learning (TL), which addresses various CV problems by reusing pre-trained models. TL requires less data and computing while delivering nearly equal accuracy, making it a prominent technique in the CV landscape. Our research focuses on TL development and how CV applications use it to solve real-world problems. We discuss recent developments, limitations, and opportunities.
\end{abstract}

\maketitle

\section{Introduction}
Storage capacity and computation power have increased considerably in recent times, especially with the expansion of the Internet and cloud services. Artificial Intelligence (AI) is one of the prominent beneficiaries of this expansion. However, even with these impressive developments, the AI models struggle with a lack of data and computation power. As a result, a company that harnesses its powers ends up in an advantageous competitive position, as vouched by 77\% of businesses in a Verizon study ~\cite{chang2018effect}. In addition, the volume of data available to a company has exploded with the increasing use of the Internet. The Internet traffic volumes increased by 30\% in 5 years, from 2017 to 2022, as reported by Nokia ~\cite{labovitz2019internet}. Abundant data and computing now empower researchers and companies to try solving complex problems that were not possible earlier. One such category is Computer Vision (CV) problems which deal with image processing ~\cite{szeliski2022computer}. Data volume is critical in Computer Vision problems using Machine Learning (ML), where an ML model learns better with more images. However, obtaining training data can be difficult and expensive for certain CV problem domains. Thus, there is an effort to reuse a trained ML model in one CV domain and apply it to a related CV domain. This effort of re-utilizing a model trained for one use case and applying it to another use case is called Transfer Learning (TL)~\cite{weiss2016survey}. Although TL can be applied to many problem domains, such as Natural Language Processing (NLP) using different techniques, such as genetic algorithms, our research paper focuses only on TL applied to CV problems. 

In the current research paper, we introduce the concepts of TL and CV, review some research papers that study these topics, and provide a gist of our research work in the background, Review of Papers, and Conclusion sections, respectively. The study aims to address recent developments in solving diverse computer vision problems through transfer learning techniques. This is not a full-fledged systematic literature review delineated by Kitchenham et al. ~\cite{kitchenham2004procedures} covering all the developmental works. Instead, we focus on addressing several popular categories that shaped the present research landscape.

\section{Background}
\begin{figure}[h]
    \centering
    \includegraphics[width=0.7\textwidth]{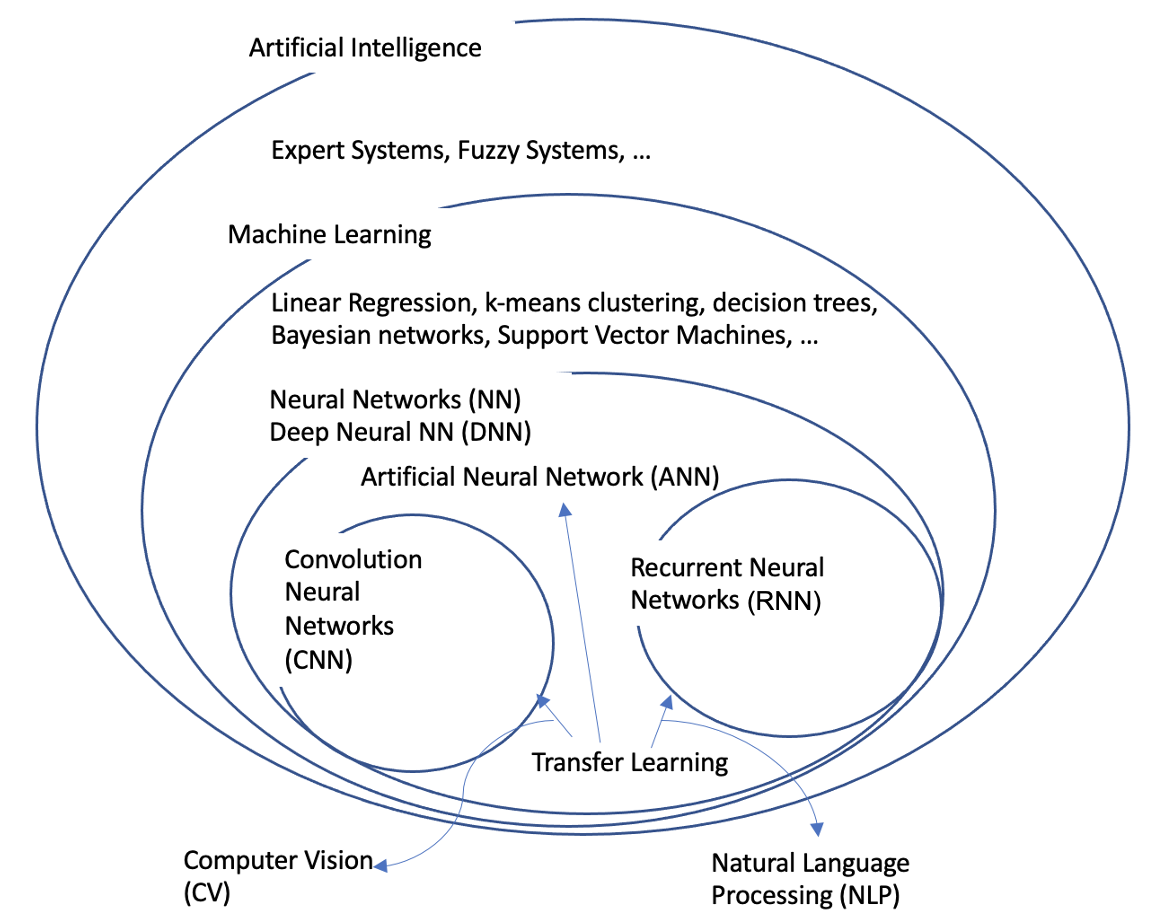}
    \caption{Venn diagram showing the relation between TL with CNN and RNN, which is a kind of DNN, which is a kind of ML, that is used in many but not all techniques of AI. Each section in the Venn diagram contains examples of the following technology.}
    \label{fig:ml-algo-rel}
\end{figure}

 In this section, we attempt to provide a preliminary understanding of the topic with relevant background. Following we outline the relationship among Artificial Intelligence (AI), Machine Learning (ML), Neural Networks (NN), Deep NN (DNN), Convolution NN (CNN), Recurrent Neural Networks (RNN), and Transfer Learning (TL). Refer to Fig. ~\ref{fig:ml-algo-rel} for enhanced reader comprehension. To maintain the scope of our research, we will limit the background section to the necessary details.

Artificial Intelligence (AI) is a field of science that utilizes machines to learn and simulate various aspects of intelligence ~\cite{dick2019artificial} to solve real-world problems. The branch is divided into rule-based Expert Systems (ES), Fuzzy Systems, and Machine Learning (ML). Expert Systems (ES) emphasizes mimicking the decision-making abilities through creating rule-based algorithms on human reasoning/logic ~\cite{russell2016artificial}. Contrary to this, fuzzy logic is a mathematical/statistical approach that deals with reasoning based on degrees of truth, rather than the traditional Boolean logic of true or false. It provides a framework for dealing with uncertainty and imprecision and more human-like decision-making in real-world problems depending upon the data. The marriage of the previous two introduces the development of Machine Learning (ML), which focuses on learning from existing data sets and making predictions/decisions through the development of algorithms and statistical models without explicit rule-based programming. There are different ML sub-fields/classifications. One such classification ~\cite{kotsiantis2006machine} is supervised learning (e.g., Linear Regression, Logistic Regression), unsupervised learning (k-means clustering), reinforcement learning (Q-learning), logic-based learning (e.g., Decision Trees), perceptron-based techniques, statistical learning algorithm (e.g., Bayesian Networks), and support vector machines (SVM).

\begin{figure}[h]
    \centering
    \includegraphics[width=0.6\textwidth]{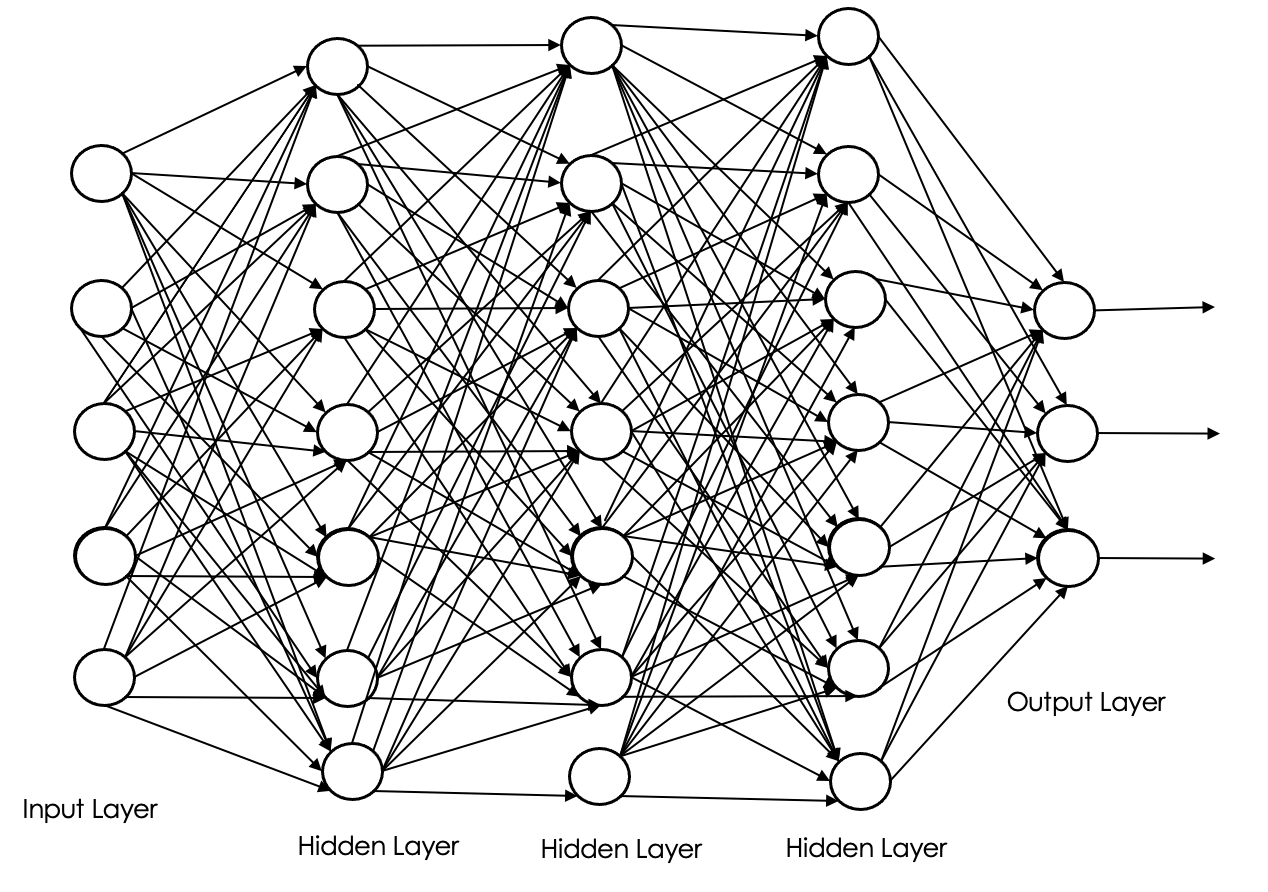}
    \caption{Diagram of a Fully Connected Deep Neural Network: The diagram illustrates the flow of an input through artificial neurons to produce an output.}
    \label{fig:dnn}
\end{figure}

Among all the machine learning techniques, neural networks (NN) is a subset that aims to imitate the interconnections of biological neurons in the human brain. They consist of interconnected nodes (neurons) arranged in layers. When activated, each neuron processes the input data and transmits it to the next layer. This process allows neural networks to learn and make decisions without being explicitly programmed ~\cite{nielsen2015neural}. In this paradigm, deep learning (DL) is a prominent approach involving the development of complex artificial neural network (ANN) architectures with multiple hidden layers to enable pattern recognition and problem-solving. Examples of deep learning architectures include Recurrent Neural Networks (RNN), Convolutional Neural Networks (CNN), and many more. Each type of neural network architecture has unique characteristics and is suitable for different data types and tasks. For example, RNNs are known for their ability to handle sequential data, while CNNs excel in computer vision tasks. These advanced neural network techniques have revolutionized many fields, including robotics\cite{neupane2024security, fernandez2024survey}, image~\cite{moore2023ura} and speech recognition~\cite{garcia2017speaker}, Natural Language Processing (NLP)~\cite{mitra2024localintel}, cybersecurity~\cite{mitra2024use, mitra2023survey}, medical diagnostics\cite{neupane2024medinsight}, etc. 

\begin{figure}[!h]
    \centering
    \includegraphics[width=1\textwidth]{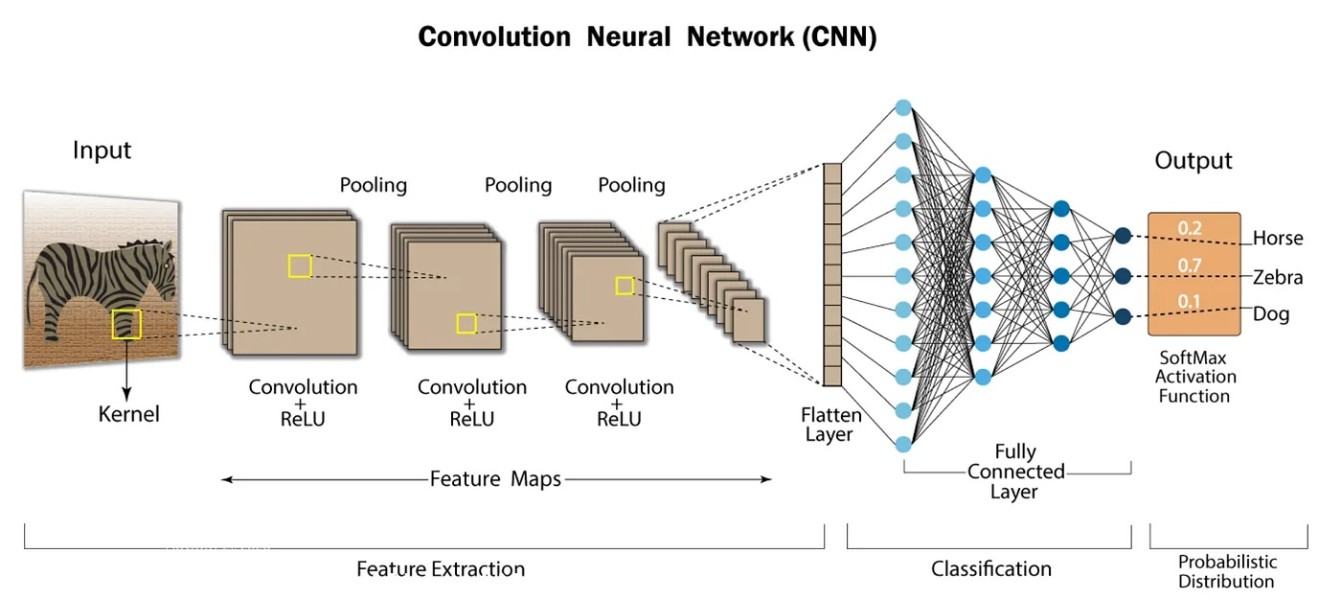}
    \caption{Convolution Neural Network \cite{cnn-linkedin}}
    \label{fig:cnn}
\end{figure}  

The Convolutional Neural Network (CNN) is an advanced architecture originally created to accurately categorize characters or zip codes in images~\cite{ibm-cnn}. This architecture has been specifically designed to significantly speed up the training and execution process for image classification and pattern recognition, making it exceptionally efficient. CNNs have found widespread applications in computer vision tasks such as facial recognition, object detection, enabling vision in robotics, and facilitating autonomous driving ~\cite{voulodimos2018deep}. A CNN typically comprises three main types of layers: convolutional layers, pooling layers, and fully connected layers, which are also referred to as fully connected neural networks (FCNs). The diagram in Figure ~\ref{fig:cnn} illustrates a CNN architecture tailored for object detection. The success of deep learning techniques depends on the perceptron, which serves as a fundamental building block of neural networks. A perceptron takes binary inputs and generates a binary output. In a perceptron-based artificial neural network, multiple perceptrons are organized in layers. They take inputs, process them to generate intermediate outputs and pass these to the next layer, leading to the final outputs. The arrangement of layers in a neural network is referred to as its architecture (CNN, RNN, Transformer, etc.). The term "deep" in deep neural network (DNN) indicates the use of multiple layers to transform input data into output, representing complex transformations, as illustrated in the accompanying Figure ~\ref{fig:dnn}.

\begin{figure}[!h]
    \centering
    \includegraphics[width=0.9\textwidth]{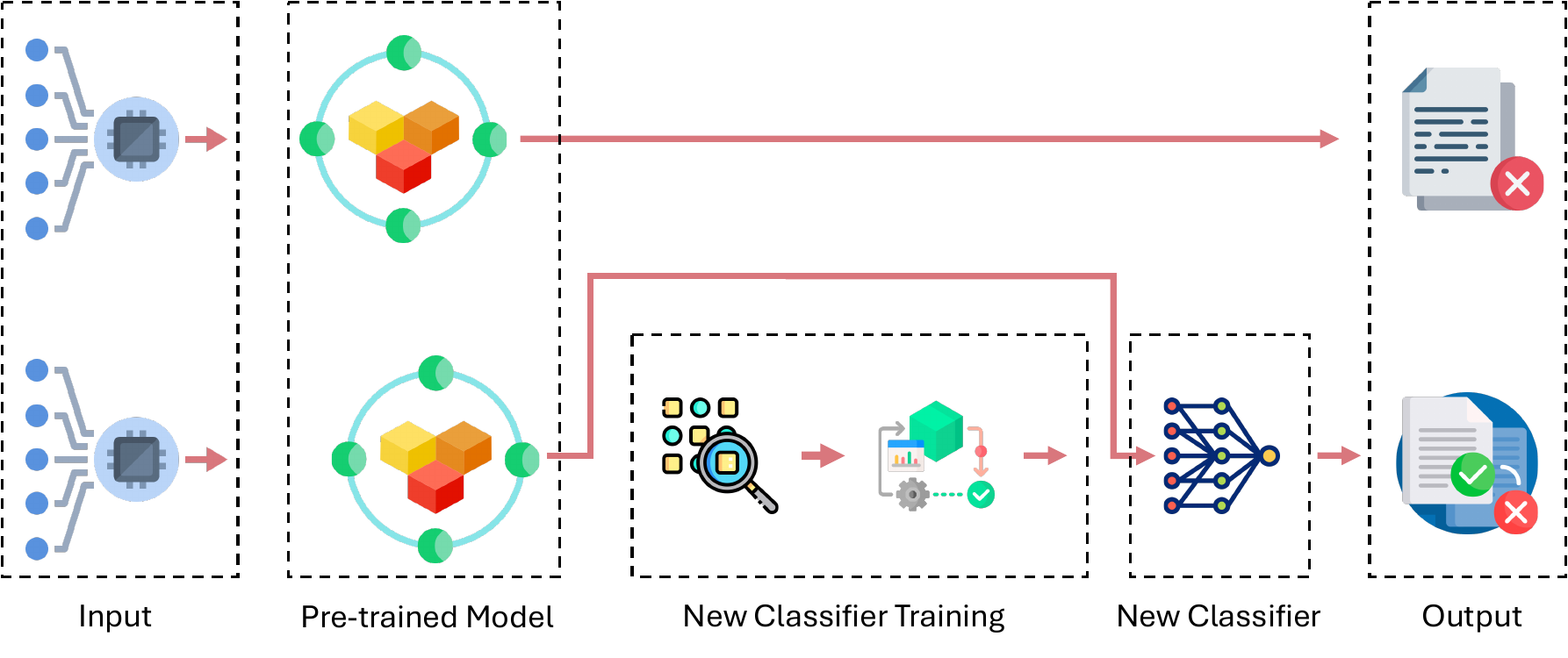}
    \caption{Transfer Learning Pipeline}
    \label{fig:transfer_learning}
\end{figure}

Transfer Learning (TL) is a machine learning technique that leverages knowledge gained from one domain to accelerate the learning process in another domain. This method is particularly valuable when it is impractical to obtain a sufficiently large dataset to train a model for a specific problem domain. For instance, in the initial stages of the COVID-19 pandemic, there was a scarcity of labeled chest X-ray data for training a network to detect the disease. However, through the application of transfer learning, researchers successfully developed a model for identifying COVID-19 ~\cite{jaiswal2021classification}. Additionally, in a study conducted by Guo et al. ~\cite{guo2019spottune}, a deep neural network model was fine-tuned using transfer learning, involving pre-training on a related task using data from the target task. TL can be applied to a variety of problem domains using different techniques. For illustration, different nature-inspired optimization algorithms like particle swarm optimization, gravitational search algorithms, charged systems search algorithms, and gray wolf optimizer algorithms are used in optimal tuning of simple Takagi-Sugeno proportional-integral fuzzy controllers involved in the position control of servo systems in ~\cite{precup2019nature}. We can adapt genetic algorithms to optimize the fuzzy logic rule base in the design phase of a fuzzy logic controller. We can leverage these optimization algorithms before the training phase and reduce learning time via transfer learning for different Neuro-Fuzzy systems. According to ~\cite{frison2003fuzzy} fuzzy logic, neural networks, and genetic algorithms can be used in improving image processing-based pattern recognition. However, we limit our study to only TL using DNN applied to CV problems. In other words, we leave the other problem domains using different techniques out of scope for the current study. The intricacies of transfer learning are comprehensively explored in the Literature Review section (Section ~\ref{section:literature_review}).

\section{Existing Research in Transfer Learning}\label{section:literature_review}
In this section, we present reviews of selected research work on the genesis of TL and its application in CV problems. The list of papers we studied in different areas are summarized in the  Table \ref{tbl:paper-reviews-summary}. 

\begin{table}[h]
    \caption{Paper review summary}
    \label{tbl:paper-reviews-summary}
    \renewcommand{\arraystretch}{1.15}
    \begin{tabularx}{1\textwidth} 
      { 
          | >{\raggedright\arraybackslash}p{1.2cm} 
          | >{\raggedright\arraybackslash}p{1.8cm} 
          | >{\raggedright\arraybackslash}p{2.8cm} 
          | >{\raggedright\arraybackslash}X | }
        \hline
            \rowcolor{lightgray} 
            \textbf{Ref.} & \textbf{Publication Year} &
            \textbf{Topic} &
            \textbf{Major Contributions}
            \\
        \hline
        ~\cite{pan2010survey},~\cite{weiss2016survey} & 2010, 2016 & What is Transfer Learning & Definition of TL, a few advantages and
TL types. \\ \hline
        ~\cite{gopalakrishnan2017deep} & 2017 & TL to find crack in pavements & Authors developed a model to automatically detect cracks in pavements using TL in a six-step method. \\ \hline
        ~\cite{kentsch2020computer} & 2020 & TL to improve tree image classification & Leverage TL to increase the quality of
the trained model with a limited dataset. \\ \hline
        ~\cite{karimi2021transfer} & 2021 & TL in medical image segmentation &  Apply TL to the
medical image segmentation issues and found the favored technique for vision applications is to pre-train a model on a source domain and fine-tune the model on a target domain. \\ \hline
        ~\cite{hridoy2021computer} & 2021 & TL to recognize Psoriasis skin disorder & TL reduces ML model training time while improving accuracy over a model built from scratch. \\ \hline
        ~\cite{deep2019leveraging} & 2019 & TL to recognize human activity & Employ TL to extract deep image features and leverage an existing pre-trained model yielded a model. \\ \hline
        ~\cite{dinh2015transfer} & 2015 & TL in Genetic Programming & Apply TL to Genetic Programming (GP) by taking final-generation individuals from the source task to the target task as first-generation individuals.\\ \hline
        
    \end{tabularx}
    
\end{table}
\newpage

\subsection{What is Transfer Learning (TL)? (Pan et al. ~\cite{pan2010survey} and Weiss et al. ~\cite{weiss2016survey})} 

\begin{wraptable}{r}{7.5cm}
\begin{center}
 \caption{Notations used to define Transfer Learning ~\cite{weiss2016survey}}
    \label{tbl:notation-table}
    \renewcommand{\arraystretch}{1.05}
\begin{tabular} { | p {1.8 cm} | p {4 cm} | }
        \hline
            \rowcolor{lightgray} 
            \textbf{Notation} &  \textbf{Description} \\
        \hline
            $X$ & Input feature space. \\
            $Y$ &Label space. \\
            $T$ &Predictive learning task. \\
            $_S$ &Denote Source. \\     
            $_T$ &Denote Target. \\
            $P(X)$ &Marginal Distribution. \\
            $P(Y|X)$ &Conditional distribution. \\
            $P(Y)$ &Label distribution. \\
            $D_S$ &Source domain data. \\
            $D_T$ &Target domain data. \\
        \hline
  
\end{tabular}
 
\end{center}
\end{wraptable}

Humans, by instinct, correlate knowledge learned from one domain and apply it to similar ones. For example, when one is familiar with a programming language, such as Java, they are already aware of concepts such as variables, data types, functions, loops, etc. They can learn relatively easily another new language, e.g., Python, by reusing the concepts they have already learned. Thus, they transferred the knowledge gathered while programming Java to learn Python. However, a computer does not inherently transfer knowledge acquired in one application to another. In this section, we study the definition of TL, with its advantages, and types.  Weiss et al. ~\cite{weiss2016survey} used the notations in Table \ref{tbl:notation-table} to define TL.

First, we define the domain and task we use to define TL. A domain $D$ consists of two parts: input feature space $X$ and a marginal probability distribution $P(X)$, where $X = \{x_1, x_2, ..., x_n\} \in X$ and $x_i$ is the $i$th feature vector and $n$ is the number of feature vectors in $X$, which is a particular learning sample in a set of all samples $X$. Thus $D = \{X, P(X)\}$. A task also consists of two parts: a label space $Y$ and an objective function $f(.)$. Thus, $T = \{Y, f(.)\}$. $f(x)$ predicts a label for a new instance $x$. $f(.)$ learns from existing learning data pair $\{x_i, y_i\} \forall i \in \{1, 2, 3, …, N\}$, where $N$ is the total number of samples. Thus $f(x) = P(y|x)$.

Second, we give an example to comprehend the notations introduced so far. If our learning task is to classify a document in a binary form, then each term is taken as a binary feature. The input space $X$ is the set of all term vectors. $x$ is a particular document or a learning sample. $x_i$ is the $i$th term vector of $x$. When we have, say, 1000 labeled documents, then we have the following data pairs for learning: $\{x_i, y_i\} \forall i \in \{1, 2, 3, …, 1000\}$, and we have $Y$ as all labels either \textit{True} or \textit{False} for all 1000 documents i.e., a vector of length 1000 of 0s or 1s and $Y_i$ represents the label of document $x_i$ at $i$th position.

Third, for simplicity, we assume that there is only one source, $D_S$, and one target domain, $D_T$ and define TL as the following: Given $D_S$, $T_S$, $D_T$ and $T_T$, TL aims to improve $fT(.)$ in $D_T$ using knowledge in $D_S$ and $T_S$, where $D_S \neq D_T$ or $T_S \neq T_T$ and $DS$ is related to $D_T$. We describe related as either explicit or implicit relationship between feature spaces of $D_S$ and $T_S$.

Fourth, in our documentation classification example, $D_S \neq D_T$ arises when $D_S$ and $D_T$ represent two different languages. $TS \neq TT$ occurs when $D_S$ and $D_T$ represent two different topics but of the same language. $T_S \neq T_T$ also implies that $Y_S \neq Y_T$ occurs or $P(Y_S|X_S) \neq P(Y_T |X_T)$, where $Y_{S_i} \in Y_S$ and $Y_{T_i} \in Y_T$. $Y_S \neq Y_T$ happens, for example, when $Y_S$ is binary and $Y_T$ is multi (e.g. 10 classes.) $P(Y_S|X_S) \neq P(Y_T |X_T)$ emerges when the source and target documents have unbalanced features.

\begin{figure}[h]
    \centering
    \includegraphics[width=1\textwidth]{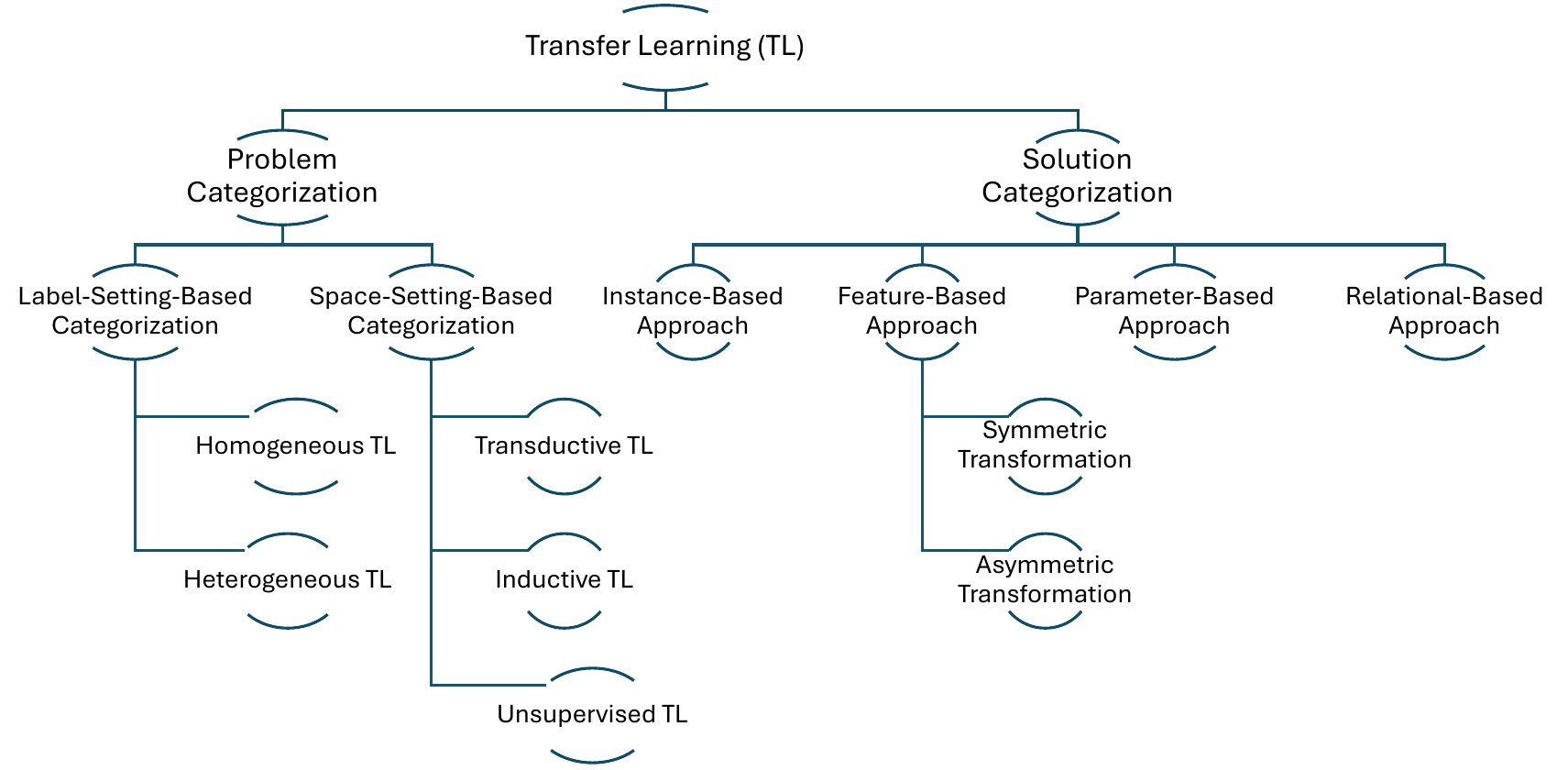}
    \caption{Transfer Learning (TL) categories}
    \label{fig:tl-categories}
\end{figure}

Fifth, the TL that we defined is using supervised learning strategy. There are other strategies for TL such as informed supervised learning ~\cite{feuz2015transfer} and semi-supervised learning ~\cite{gong2012geodesic}. Moreover, TL has many categories such as Inductive, Unsupervised, and Transductive using different approaches based on what to transfer, how to transfer, and when to transfer. In addition, TL could also be categorized based on different learning approaches e.g., instance-based, parameter-based, etc. ~\cite{zhuang2020comprehensive} as shown in Fig. ~\ref{fig:tl-categories}.

Finally, we highlight the points below, which are critical to apply successfully TL as gathered by the authors in ~\cite{neyshabur2020being}:
\begin{itemize}
    \item Feature reuse is key to a successful transfer. The authors concluded so after using a set of shuffled small blocks of the images to train and compare the model outputs. Thus, $Y_S$ should be similar to $X_S$.

    \item Preserving the data distribution is also essential for a successful transfer. The authors experimented and reasoned so after experimenting with different image block sizes and image channels. They call it 'low-level statistics.'

    \item Model mistakes on the source domain, $D_S$, also carry over to the target domain $D_T$. Thus, $E_S$ and $E_T$ are similar, where $E$ is the model error.

    \item Lower layers of the model architecture (neural network) capture the general features, which are reused in the transfer process.

    \item Higher layers of the model are more impacted by the parameters and hence need to be replaced by the target domain parameters.

    \item Model training resumption for an earlier training check-point does not reduce the accuracy.
\end{itemize}

\subsection{TL to find crack in pavements (Khaitan et al. ~\cite{gopalakrishnan2017deep})}

\begin{figure}
    \centering
    \includegraphics[width=\textwidth]{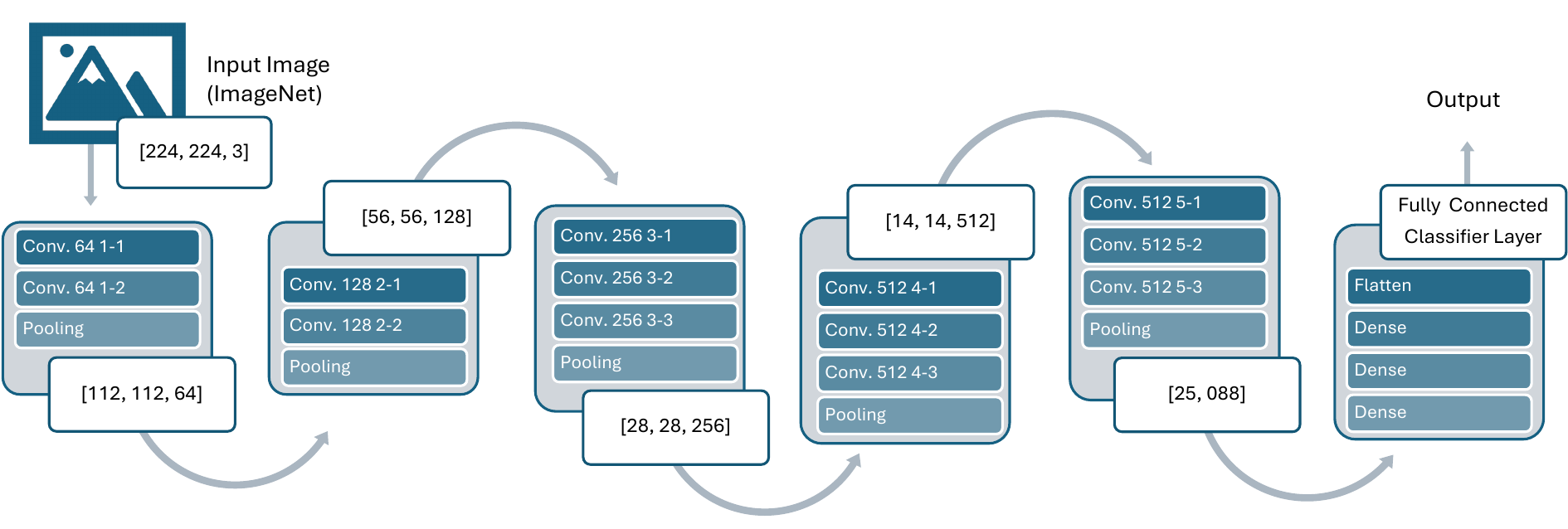}
    \caption{VGG-16 CNN schematic}
    \label{fig:khaitan-pave-crack}
\end{figure}

The authors~\cite{gopalakrishnan2017deep} conducted their work based on the idea that using deep learning-trained models on large scale image datasets such as ImageNet ~\cite{krizhevsky2017imagenet} and \textit{transferred} their ability to a new classification on a new dataset, from another domain, which is cheaper and more efficient than training the model from scratch ~\cite{bar2015chest}. Therefore, they truncated a VGG-16 model that is developed by the Visual Geometry Group (VGG) at the University of Oxford using the ImageNet dataset. However, they only removed the last fully connected classifier layers, as shown in Fig. ~\ref{fig:khaitan-pave-crack}. They used the truncated model with their input images to produce a new binary classifier model that predicts if an image has a crack or not.

Using this approach, the authors developed a model to automatically detect cracks in pavements using TL in a six-step method. First, they pre-processed the raw pavement images sampled from a publicly available dataset ~\cite{elkins2018long}. They also included images with lane markings, oil spills, etc., so the model generalizes better for real-world pictures. In addition, they split the labeled dataset into training, validation, and testing. Second, they run the labeled pavement images through the truncated VGG-16 model to generate new images and extract new features. Third, they use the generated image to train a binary classifier. Fourth, they used different ML techniques, such as NN, Random Forest (RF), etc., to compare the model performance. Fourth, they used the standard model training procedure to train, test, and validate using the corresponding datasets. Finally, they report the model training metrics in Confusion Matrix and its derived metrics such as accuracy, precision, recall, etc. The research concluded that the binary classifier model using the NN performed best. For the NN, they used the 'Adam' optimizer, 'Rectified Linear-unit (ReLu)' activation function in the hidden layers, 'Softmax' in the output layer, 256 neurons in the hidden layer, 32 batch size, 50 epochs, and Mean-Squared-Error (MSE) loss function. However, these hyperparameters are specific to their image set data and should be changed to create a model for another domain with a different image set. But broadly, these parameters can be used as initial values to start training the model.

\subsection{TL to improve tree image classification (Caceres et al. ~\cite{kentsch2020computer})}

Caceres et al.~\cite{kentsch2020computer} stated that they leveraged TL to increase the quality of the trained model with a limited dataset. They detected invasive trees among the classified trees in the forest images taken by drones. However, they only could gather a limited dataset because of the hardship of collecting data due to natural phenomena such as inclement weather (e.g., rain.) Moreover, they concluded that TL was critical in building an ML model that predicts accurately for their domain. In addition, they concluded that using a data set similar to their dataset in the TL process improved the model accuracy. 

The researchers followed a five-step approach:

\begin{enumerate}
    \item They developed a model reusing the ResNet50 ~\cite{he2016deep} architecture to classify tree species using the Multi-Label Patch algorithm.

    \item They analyzed the model quality by testing with the limited data set. 

    \item They studied the learning rate impact on the ML model by adopting three approaches, namely, (a) without using TL, (b) using TL once with ImageNet ~\cite{krizhevsky2017imagenet}, and (c) using TL twice: once with Planet Satellite ~\cite{ching2019understanding} data set and second with the domain dataset. 

    \item They improved the accuracy and the learning time (by reducing the computation) of the ML model, tuning it by the conclusions from the analysis.

    \item They applied the model to detect invasive tree species in a coastal forest.
\end{enumerate}

The authors used the following labels: deciduous, evergreen, uncovered, river, and man-made. In addition, they considered frozen and unfrozen forms in their three approaches in the third step. Only the final layer ML architecture is changed in the frozen form, whereas all the layers are changed in the unfrozen form. Both forms are used in all three approaches for model training. Moreover, the starting for corresponding to the approaches are:
\begin{itemize}
    \item \textbf{Random}: In this approach, all weights were initialized to random weights. The approach intended to study if TL is necessary at all. However, only the unfrozen form experiment testbed was included in the comparison analysis because the frozen form yielded bad results.

    \item \textbf{ResNet architecture with ImageNet dataset}: The approaches initialized ResNet50 ML weights to the preloaded ImageNet weights. The study is intended to discover if a general-purpose image dataset could be used for this application domain. Thus, two test experiment testbeds were created, namely, RN50+F and RN50+UN, corresponding to ResNet with frozen form and ResNet with unfrozen form.

    \item \textbf{ResNet architecture with Planet Images dataset}: The approaches initialized ResNet-50 ML weights to the pre-loaded Planet weights. The analysis was to discover if a related image dataset could be used to create a base model for their problem domain. Furthermore, the investigators used the frozen models to train on their domain dataset. Thus, four test-beds were created: ResNet with frozen and unfrozen form, and finally, domain-specific ML model with frozen and unfrozen form.
\end{itemize}

\begin{figure}[h]
    \centering
    \includegraphics[width=0.65\textwidth]{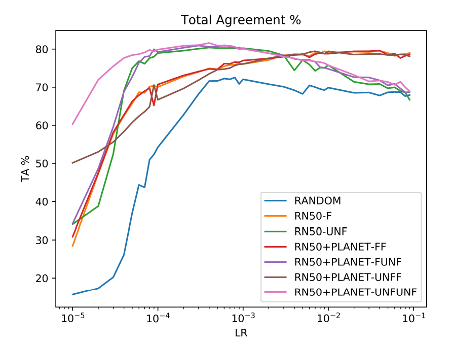}
    \caption{Multiple TLs with frozen and unfrozen forms. ~\cite{kentsch2020computer}}
    \label{fig:kentsch-tree}
\end{figure}

Their experiment used various learning rates ranging from $10^{-5}$ to $0.9$ and gathered empirical data from the experimental testbed settings. Fig. ~\ref{fig:kentsch-tree} shows the Total Agreement (FA) metric graph with various learning rates for the five experimental testbeds. TA is the percentage of correct predicted labels, i.e., correct predicted divided by the total samples.

The researchers concluded from the experiment that the ResNet with Planet data using unfrozen from as a starting model applied to domain data set in the unfrozen architecture gives the best result. 

\subsection{TL in medical image segmentation (Karimi et al. ~\cite{karimi2021transfer})}

In medical image segmentation research Karimi et al.~\cite{karimi2021transfer} indicated that although there are different TL techniques, such as Deep Adaption Networks ~\cite{long2015learning} and few/zero-shot learning ~\cite{zhang2015zero}, the favored technique for vision applications is to pre-train a model on a source domain and fine-tune the model on a target domain. It is the same approach that the other research papers have employed that we have reviewed.

The authors made the following observations:
\begin{itemize}
    \item TL reduces the training time on the target task, $T$ of target domain, $D_T$. Moreover, the accuracy depends on the data, and the accuracy improvement is marginal. 

    \item The model parameters used in the TL process remain relatively the same, regardless of using either randomly-initialized or pre-trained values. 
    
    \item The convolution filters remain almost unchanged during the TL training phase. 
    
    \item The model accuracy increases by freezing the network encoder section and training only the decoder section.
    
    \item TLs produce different FCN representations than the FCNs with random initialization. However, the variability in the FCN set by TL can be as high as the FCN set by random initialization.
    
    \item Feature reuse can be more significant in the deeper layer than in the earlier encoding layers.
    
    \item Applying TL to different model architectures yields similar results.

\end{itemize}

\begin{figure}[h]
    \centering
    \includegraphics[width=0.65\textwidth]{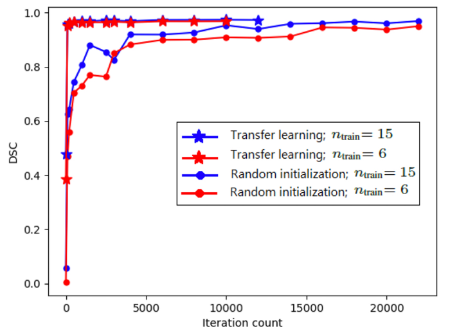}
    \caption{Multiple TLs with frozen and unfrozen forms. ~\cite{karimi2021transfer}}
    \label{fig:karimi-medicalimage}
\end{figure} 
The researchers made these observations by applying TL to the medical image segmentation issues. They leveraged 16 datasets containing 11 body parts for the model training. Next, they preprocessed the data by resampling and normalizing the datasets. Finally, they used 70\% of the combined dataset for model training and validation and the rest for model training. Their experiment used four networks, HRNet~\cite{wang2020deep}, UNet++~\cite{zhou2018unet++}, Tiramisu~\cite{jegou2017one} (a modified DenseNet~\cite{jaiswal2021classification},) and Autofocus~\cite{qin2018autofocus}, as is, and a modified V-Net~\cite{milletari2016v} version to deduce those observations. The modified version has six layers. It also has three encoding and three decoding layers. They used the 'Adam' optimizer with an initial learning rate of $10^{-4}$, which was reduced by $0.9$ after every 2000 training iterations if the loss did not decrease. In addition, they initialized the convolution filters with zero-mean and $\sqrt{2/n}$, where $n$ is the number of connections to the convolutional filter from the previous layer, Gaussian random variables. Moreover, they kept kernel size to 3 and used the 'ReLU' activation function after convolutional operations. In addition, they gathered the following metrics from the experiments: Dice Similarity Coefficient (DSC), Average Symmetric Surface Distance (ASSD), the 95 percentile of the Hausdorff Distance (HD95), and the F1-score. The Fig. ~\ref{fig:karimi-medicalimage} shows DSC against ML training iteration count for the liver organ body part trained using TL and from scratch (authors also called it 'random initialization). The authors used three MRI datasets (out of 16) to train the model and then fine-tune it with another Liver-CT dataset (out of 16.) The Fig. ~\ref{fig:karimi-medicalimage} shows that the model convergence is faster than the one trained from scratch. However, the model accuracy between the two methods is not significant. Moreover, this observation was the same for all five network architectures.

\subsection{TL to recognize Psoriasis skin disorder (Hridoy et al. ~\cite{hridoy2021computer})}

Hridoy et al.~\cite{hridoy2021computer} concluded study to recognize Psoriasis skin disorder and found that TL reduces ML model training time while improving accuracy over a model built from scratch. They reused EfficientNet pre-trained ML models ~\cite{tan2019efficientnet} to reclassify a skin order dataset to recognize a Psoriasis type from 12 classes. EfficientNet models are trained on ImageNet ~\cite{krizhevsky2017imagenet}, which has more than 1.2 million images. Their methodology followed the steps mentioned below:
\begin{enumerate}

   \item They gathered and labeled 6,300 sample data.

    \item They synthesize the sample data to augment their original image set to 52,500. They used image transformation techniques such as rotation and horizontal and vertical flips.

    \item They pre-processed sample data and reshaped the sample images into eight different dimensions.

    \item They selected all 8 EfficientNet models. All layers used 'Swish' as the action function except the last fully connected layer, which used 'softmax.' The output layer has 1000 inputs and 21 outputs that match the 21 Psoriasis classes. The authors used the 'early stopping' method while training the NN. They also used 'adam' optimizer. They used 42,00 samples to train, 4,200 to validate, and 6,300 to test.

    \item They trained the NN and gathered the ML performance confusion metrics. They calculated the sensitivity, specificity, accuracy, and precision metrics.

    \item They concluded that the ML model accuracy was 97.1\% which is better than the accuracy of several other ML models, as shown in the Table \ref{tbl:hridoy-ml-performance}.
    
\end{enumerate}

\begin{table}
    \caption{TL using EfficientNet vs others ~\cite{hridoy2021computer}}
        \label{tbl:hridoy-ml-performance}
    \renewcommand{\arraystretch}{1.15}
    \begin{tabularx}{\textwidth} { 
          | >{\raggedright\arraybackslash}p{5cm} 
          | >{\raggedright\arraybackslash}p{3.5cm} 
          | >{\raggedleft\arraybackslash}p{1.5cm} 
          | >{\raggedleft\arraybackslash}X | }
        \hline
            \rowcolor{lightgray} 
            \textbf{Study} &                 \textbf{Method} &
            \textbf{Classes} &
            \textbf{Accuracy (\%)}
            \\
        \hline
        Nurul Akmalia et al. ~\cite{akmalia2019skin} & LBP, CNN & 6 & 92 \\ \hline
        Rola EL SALEH et al. ~\cite{el2019deep} & AlexNet & 10 & 88 \\ \hline
        Evgin Goceri et al. ~\cite{goceri2019analysis} & ResNet & 6 & 97.01 \\ \hline
        Jainesh Rahtod et al. ~\cite{rathod2018diagnosis} & CNN & 5 & 70 \\ \hline
        Nazia Hameed et al. ~\cite{shanthi2020automatic} & SVM, AlexNet & 5 & 86.21 \\ \hline
        Shuchi Bhadula et al.~\cite{bhadula2019machine} & CNN & 3 & 96 \\ \hline
        Tamanna T. K. Munia et al. ~\cite{munia2017automatic} & K-means clustering & 3 & 93.83 \\ \hline
        Md. Aminur Rab Ratul et al. ~\cite{ratul2020skin} & Inception-V3 & 7 & 89.81 \\ \hline
        Current study ~\cite{hridoy2021computer} & EfficientNet-B7 & 21 & 97.1 \\ \hline            
    \end{tabularx}
    
\end{table}

\subsection{TL to recognize human activity (Deep et al. ~\cite{deep2019leveraging})}
The authors in ~\cite{deep2019leveraging} concluded that employing TL to extract deep image features and leverage an existing pre-trained model yielded a model with 96.95\% accuracy. Moreover, they mentioned that the accuracy of the model using TL performs 1 - 6\% better than the results achieved from other approaches. However, they alerted that their technique might be flawed because they are using pre-trained ImageNet ~\cite{krizhevsky2017imagenet} weights containing images of several categories. The trained ML model classifies seven human activities: bend, back, jump, run, side, skip, walk, wave1 and wave2. First, they choose the Weizmann dataset ~\cite{gorelick2007actions} for their ML model, which consists of 9 people performing ten activities. Second, they pre-processed the video and extracted individual frames from the videos. They collected 4,917 frames for all seven activities. They used 70\% of the frames to train, 10\% to validate, and 10\% to test the ML model. Third, they experimented with three CNNs, namely, VGG-16, VGG-19, and Google's InceptionNet-v3. Thus, they leveraged the knowledge from the large-scale ImageNet dataset. They extracted the features from the penultimate CNN layers. Finally, they created a confusion matrix and using it, they calculated the accuracy, precision, recall, and f1-score for the three CNNs, as shown in the Table~\ref{tbl:deep-recognize-human}.

\begin{table}
    \caption{Results for 3 CNNs ~\cite{deep2019leveraging}}
    \label{tbl:deep-recognize-human}
    \begin{tabularx}{0.85\textwidth} { 
      | >{\centering\arraybackslash}p{0.15 \textwidth}|X
      | >{\centering\arraybackslash}X  
      | >{\centering\arraybackslash}X    
      | >{\centering\arraybackslash}X    
      | >{\centering\arraybackslash}X |     
      }
        \hline
        \rowcolor{lightgray} 
            \textbf{Model} &                 
            \textbf{Accuracy (\%)} &
            \textbf{Precision (\%)} &
            \textbf{Recall (\%)} & 
            \textbf{F1-score (\%)}
            \\
        \hline
        VGG-16 & 96.95 & 97.00 & 97.00 & 97.00 \\ \hline
        VGG-19 & 96.54 & 97.00 & 97.00 & 97.00 \\ \hline
        Inception-v3 & 95.63 & 96.00 & 96.00 & 96.00 \\ \hline        
    \end{tabularx}
    
\end{table}

\subsection{TL in Genetic Programming (Thi Thu Huong Dinh et al. ~\cite{dinh2015transfer})}

Thi Thu Huong Dinh et al.~\cite{dinh2015transfer} applied TL to Genetic Programming (GP) by taking final-generation individuals, or
algorithmic solutions, from the source task to the target task as first-generation individuals. They
compared the results of the TL-assisted GP and GP without any help from TL, which is called standard
GP. They found that it improved training errors, especially with unseen data, compared to standard GP.
Furthermore, they claimed that limiting the transferred individuals reduced unnecessary code growth,
also called code bloat. Thus, the technique helped them to avoid negative transfer, or when transferring
does not aid the GP’s performance. They used a parameter-based GP model which shared some source and target
parameters and used different configurations of k, where k is the variable of the parametric values, either expressing a percentage of individuals or the number of individuals transferred. The
three used TL methods are following:

\begin{enumerate}

\item \textbf{FullTree}: k\% of best individuals from the last generation of the source problem are
transferred to the first generation of the target problem.

\item \textbf{SubTree}: From k\% of best individuals, a random subtree is chosen from each last-
generation individual of the source problem, pooled, and transferred to the target
problem as first-generation individuals.

\item \textbf{BestGen}: k best individuals from each generation of the evolutionary process are taken,
pooled, and transferred like the other methods.

\end{enumerate}

The experiment was conducted over two families of regression problems: polynomial and trigonometric. The
target and source problems are similar, but the target problem is more complex. For Full-Tree and
Sub-Tree, the authors used k values of 25, 50, 75, and 100. With BestGen, they had two experiments. The
first is the non-restricted experiment, where they transferred k best individuals—in this case, k being the
number of individuals; k = 1, 2—from each generation of the source problem to the target problem, with
no limit on the number of nodes that make up the individuals. The second is the restricted experiment,
where they restricted the transferred individuals’ sizes to a maximum of 50 nodes but followed the
same procedure as the first experiment. Finally, the authors concluded the experiment with three tests: training error, GP’s generalization
ability, and code bloat.

\begin{enumerate}

\item \textbf{Training Error}: FullTree performed its best at k=25. However, compared to standard GP,
it did not do well with trigonometric problems. SubTree and BestGen both had small
training errors with SubTree performing its best at k=25 and 50. BestGen (restricted and
non-restricted) performed consistently with all four configurations.

\item \textbf{Generalization Ability}: TL-assisted GP was better than standard GP overall. Their testing
errors, especially with SubTree and BestGen, had been significantly better when faced
with unseen data.

\item \textbf{Code Bloat}: FullTree incurred higher code growth than standard GP, showing negative
transfer at times. SubTree and the restricted BestGen, however, bloated less than
standard GP. Thus, restricting the transferred individuals’ sizes helped TL reduce code
growth.

\end{enumerate}

In the following [Section \ref{section:discussion}], we will conclude our study by briefly summarizing existing research and our findings from both challenges and opportunities perspectives.    

\section{Discussion}\label{section:discussion}
Although CV problems have existed for a while, recent data and compute availability have sparked interest in revisiting them. However, obtaining sufficient data to train an ML model from scratch remains a challenge, as an ML model for CV requires many examples for training. Gathering enough data may not be possible in specific domains. For illustration, there might not be enough image scans available for a relatively newly discovered disease in health care. Similarly, there are often financial or time constraints when using nearly infinite computing, especially with cloud computing. Developing a usable ML model with a shorter training time is cost-effective. Researchers and academicians have proposed several techniques to accomplish this by using proper hyperparameters, such as early training stopping when the algorithm learning drops below a certain threshold \cite{bischl2023hyperparameter}. 

Hence, considering limited data availability and efficient compute usage, one needs to train usable ML models. One method to do so is the transfer learning. In this survey paper, we studied several computer vision problems that leverage TL to solve them. Khaitan et al. \cite{gopalakrishnan2017deep} concluded that training a base model already trained on large-scale image data with a limited pavement crack data proved cheaper. Caceres et al. \cite{kentsch2020computer} had to use TL to train a model from a base model as they had limited evasive plant data. Karimi et al. \cite{karimi2021transfer} have noticed significantly less training time and increased model accuracy when they used TL to train on limited medical image data. Similar conclusions were drawn by Hridoy et al. \cite{hridoy2021computer}, where they had only 6,000 samples of psoriasis skin disorder to train their model. Deep et al. \cite{deep2019leveraging} also demonstrated that using TL produced a model with up to 6\% better performance than other approaches. Hence, the TL method is a proven technique to train a model with limited data for the CV domain. The method also has a benign advantage: cheaper.  

\section{Conclusion}
In this paper, we reviewed Transfer Learning, an ML model training optimization technique applied to the Computer Vision problems. TL re-purposes a pre-trained model on a similar problem domain with a different task. TL also requires fewer samples than a model trained from scratch does. It reuses (freezes) most of the layers of the trained NN architecture and only retrains the final fully connected classifiers layers with fewer samples. TL has shown better training speed because of training only a few layers and, at times, better accuracy. For instance, training an ML model with fewer samples achieves a similar performance than is trained on the entire set, as concluded empirically by the authors in the study ~\cite{huh2016makes}. Finally, although we have studied only optimizing training using TL in computer vision, TL can be applied to many other domains. A few examples of such domains are autonomous driving, gaming, healthcare, spam filtering, speech, and natural language processing applications.


\bibliographystyle{ACM-Reference-Format}
\bibliography{sample-base}


\begin{thebibliography}{59}


\ifx \showCODEN    \undefined \def \showCODEN     #1{\unskip}     \fi
\ifx \showDOI      \undefined \def \showDOI       #1{#1}\fi
\ifx \showISBNx    \undefined \def \showISBNx     #1{\unskip}     \fi
\ifx \showISBNxiii \undefined \def \showISBNxiii  #1{\unskip}     \fi
\ifx \showISSN     \undefined \def \showISSN      #1{\unskip}     \fi
\ifx \showLCCN     \undefined \def \showLCCN      #1{\unskip}     \fi
\ifx \shownote     \undefined \def \shownote      #1{#1}          \fi
\ifx \showarticletitle \undefined \def \showarticletitle #1{#1}   \fi
\ifx \showURL      \undefined \def \showURL       {\relax}        \fi
\providecommand\bibfield[2]{#2}
\providecommand\bibinfo[2]{#2}
\providecommand\natexlab[1]{#1}
\providecommand\showeprint[2][]{arXiv:#2}

\bibitem[Akmalia et~al\mbox{.}(2019)]%
        {akmalia2019skin}
\bibfield{author}{\bibinfo{person}{Nurul Akmalia}, \bibinfo{person}{Poltak Sihombing}, {et~al\mbox{.}}} \bibinfo{year}{2019}\natexlab{}.
\newblock \showarticletitle{Skin Diseases Classification Using Local Binary Pattern and Convolutional Neural Network}. In \bibinfo{booktitle}{\emph{2019 3rd International Conference on Electrical, Telecommunication and Computer Engineering (ELTICOM)}}. IEEE, \bibinfo{pages}{168--173}.
\newblock


\bibitem[Bar et~al\mbox{.}(2015)]%
        {bar2015chest}
\bibfield{author}{\bibinfo{person}{Yaniv Bar}, \bibinfo{person}{Idit Diamant}, \bibinfo{person}{Lior Wolf}, \bibinfo{person}{Sivan Lieberman}, \bibinfo{person}{Eli Konen}, {and} \bibinfo{person}{Hayit Greenspan}.} \bibinfo{year}{2015}\natexlab{}.
\newblock \showarticletitle{Chest pathology detection using deep learning with non-medical training}. In \bibinfo{booktitle}{\emph{2015 IEEE 12th international symposium on biomedical imaging (ISBI)}}. IEEE, \bibinfo{pages}{294--297}.
\newblock


\bibitem[Bhadula et~al\mbox{.}(2019)]%
        {bhadula2019machine}
\bibfield{author}{\bibinfo{person}{Shuchi Bhadula}, \bibinfo{person}{Sachin Sharma}, \bibinfo{person}{Piyush Juyal}, {and} \bibinfo{person}{Chitransh Kulshrestha}.} \bibinfo{year}{2019}\natexlab{}.
\newblock \showarticletitle{Machine learning algorithms based skin disease detection}.
\newblock \bibinfo{journal}{\emph{International Journal of Innovative Technology and Exploring Engineering (IJITEE)}} \bibinfo{volume}{9}, \bibinfo{number}{2} (\bibinfo{year}{2019}), \bibinfo{pages}{4044--4049}.
\newblock


\bibitem[Bischl et~al\mbox{.}(2023)]%
        {bischl2023hyperparameter}
\bibfield{author}{\bibinfo{person}{Bernd Bischl}, \bibinfo{person}{Martin Binder}, \bibinfo{person}{Michel Lang}, \bibinfo{person}{Tobias Pielok}, \bibinfo{person}{Jakob Richter}, \bibinfo{person}{Stefan Coors}, \bibinfo{person}{Janek Thomas}, \bibinfo{person}{Theresa Ullmann}, \bibinfo{person}{Marc Becker}, \bibinfo{person}{Anne-Laure Boulesteix}, {et~al\mbox{.}}} \bibinfo{year}{2023}\natexlab{}.
\newblock \showarticletitle{Hyperparameter optimization: Foundations, algorithms, best practices, and open challenges}.
\newblock \bibinfo{journal}{\emph{Wiley Interdisciplinary Reviews: Data Mining and Knowledge Discovery}} \bibinfo{volume}{13}, \bibinfo{number}{2} (\bibinfo{year}{2023}), \bibinfo{pages}{e1484}.
\newblock


\bibitem[Chang et~al\mbox{.}(2018)]%
        {chang2018effect}
\bibfield{author}{\bibinfo{person}{Younghoon Chang}, \bibinfo{person}{Siew~Fan Wong}, \bibinfo{person}{Uchenna Eze}, {and} \bibinfo{person}{Hwansoo Lee}.} \bibinfo{year}{2018}\natexlab{}.
\newblock \showarticletitle{The effect of IT ambidexterity and cloud computing absorptive capacity on competitive advantage}.
\newblock \bibinfo{journal}{\emph{Industrial Management \& Data Systems}} (\bibinfo{year}{2018}).
\newblock


\bibitem[Ching et~al\mbox{.}(2019)]%
        {ching2019understanding}
\bibfield{author}{\bibinfo{person}{Derek Ching}, \bibinfo{person}{Yuetian Li}, {and} \bibinfo{person}{Guanzhou Song}.} \bibinfo{year}{2019}\natexlab{}.
\newblock \showarticletitle{Understanding the Amazon from Space}.
\newblock \bibinfo{journal}{\emph{Northeastern Univ., Boston, MA, USA, Tech. Rep}} (\bibinfo{year}{2019}).
\newblock


\bibitem[Deep and Zheng(2019)]%
        {deep2019leveraging}
\bibfield{author}{\bibinfo{person}{Samundra Deep} {and} \bibinfo{person}{Xi Zheng}.} \bibinfo{year}{2019}\natexlab{}.
\newblock \showarticletitle{Leveraging CNN and transfer learning for vision-based human activity recognition}. In \bibinfo{booktitle}{\emph{2019 29th International Telecommunication Networks and Applications Conference (ITNAC)}}. IEEE, \bibinfo{pages}{1--4}.
\newblock


\bibitem[Dick(2019)]%
        {dick2019artificial}
\bibfield{author}{\bibinfo{person}{Stephanie Dick}.} \bibinfo{year}{2019}\natexlab{}.
\newblock \showarticletitle{Artificial intelligence}.
\newblock  (\bibinfo{year}{2019}).
\newblock


\bibitem[Dinh et~al\mbox{.}(2015)]%
        {dinh2015transfer}
\bibfield{author}{\bibinfo{person}{Thi Thu~Huong Dinh}, \bibinfo{person}{Thi~Huong Chu}, {and} \bibinfo{person}{Quang~Uy Nguyen}.} \bibinfo{year}{2015}\natexlab{}.
\newblock \showarticletitle{Transfer learning in genetic programming}. In \bibinfo{booktitle}{\emph{2015 IEEE Congress on Evolutionary Computation (CEC)}}. IEEE, \bibinfo{pages}{1145--1151}.
\newblock


\bibitem[EL~SALEH et~al\mbox{.}(2019)]%
        {el2019deep}
\bibfield{author}{\bibinfo{person}{Rola EL~SALEH}, \bibinfo{person}{Sambit BAKHSHI}, {and} \bibinfo{person}{NAIT-ALI Amine}.} \bibinfo{year}{2019}\natexlab{}.
\newblock \showarticletitle{Deep convolutional neural network for face skin diseases identification}. In \bibinfo{booktitle}{\emph{2019 Fifth International Conference on Advances in Biomedical Engineering (ICABME)}}. IEEE, \bibinfo{pages}{1--4}.
\newblock


\bibitem[Elkins et~al\mbox{.}(2018)]%
        {elkins2018long}
\bibfield{author}{\bibinfo{person}{Gary~E Elkins}, \bibinfo{person}{Peter~N Schmalzer}, \bibinfo{person}{Travis Thompson}, \bibinfo{person}{Amy Simpson}, {et~al\mbox{.}}} \bibinfo{year}{2018}\natexlab{}.
\newblock \bibinfo{booktitle}{\emph{Long-term pavement performance information management system: Pavement performance database user reference guide}}.
\newblock \bibinfo{type}{{T}echnical {R}eport}. \bibinfo{institution}{Turner-Fairbank Highway Research Center}.
\newblock
\newblock
\shownote{Accessed: 2023-01-16}.


\bibitem[Fernandez et~al\mbox{.}(2024)]%
        {fernandez2024survey}
\bibfield{author}{\bibinfo{person}{Ivan~A Fernandez}, \bibinfo{person}{Subash Neupane}, \bibinfo{person}{Trisha Chakraborty}, \bibinfo{person}{Shaswata Mitra}, \bibinfo{person}{Sudip Mittal}, \bibinfo{person}{Nisha Pillai}, \bibinfo{person}{Jingdao Chen}, {and} \bibinfo{person}{Shahram Rahimi}.} \bibinfo{year}{2024}\natexlab{}.
\newblock \showarticletitle{A Survey on Privacy Attacks Against Digital Twin Systems in AI-Robotics}.
\newblock \bibinfo{journal}{\emph{arXiv preprint arXiv:2406.18812}} (\bibinfo{year}{2024}).
\newblock


\bibitem[Feuz and Cook(2015)]%
        {feuz2015transfer}
\bibfield{author}{\bibinfo{person}{Kyle~D Feuz} {and} \bibinfo{person}{Diane~J Cook}.} \bibinfo{year}{2015}\natexlab{}.
\newblock \showarticletitle{Transfer learning across feature-rich heterogeneous feature spaces via feature-space remapping (FSR)}.
\newblock \bibinfo{journal}{\emph{ACM transactions on intelligent systems and technology (TIST)}} \bibinfo{volume}{6}, \bibinfo{number}{1} (\bibinfo{year}{2015}), \bibinfo{pages}{1--27}.
\newblock


\bibitem[Fris{\'o}n and Masip(2003)]%
        {frison2003fuzzy}
\bibfield{author}{\bibinfo{person}{Pilar~Sobrevilla Fris{\'o}n} {and} \bibinfo{person}{Eduard~Montseny Masip}.} \bibinfo{year}{2003}\natexlab{}.
\newblock \showarticletitle{Fuzzy sets in computer vision: an overview}.
\newblock \bibinfo{journal}{\emph{Mathware \& soft computing}} (\bibinfo{year}{2003}).
\newblock


\bibitem[Garcia-Romero et~al\mbox{.}(2017)]%
        {garcia2017speaker}
\bibfield{author}{\bibinfo{person}{Daniel Garcia-Romero}, \bibinfo{person}{David Snyder}, \bibinfo{person}{Gregory Sell}, \bibinfo{person}{Daniel Povey}, {and} \bibinfo{person}{Alan McCree}.} \bibinfo{year}{2017}\natexlab{}.
\newblock \showarticletitle{Speaker diarization using deep neural network embeddings}. In \bibinfo{booktitle}{\emph{2017 IEEE International Conference on Acoustics, Speech and Signal Processing (ICASSP)}}. IEEE, \bibinfo{pages}{4930--4934}.
\newblock


\bibitem[Goceri(2019)]%
        {goceri2019analysis}
\bibfield{author}{\bibinfo{person}{Evgin Goceri}.} \bibinfo{year}{2019}\natexlab{}.
\newblock \showarticletitle{Analysis of deep networks with residual blocks and different activation functions: classification of skin diseases}. In \bibinfo{booktitle}{\emph{2019 Ninth international conference on image processing theory, tools and applications (IPTA)}}. IEEE, \bibinfo{pages}{1--6}.
\newblock


\bibitem[Gong et~al\mbox{.}(2012)]%
        {gong2012geodesic}
\bibfield{author}{\bibinfo{person}{Boqing Gong}, \bibinfo{person}{Yuan Shi}, \bibinfo{person}{Fei Sha}, {and} \bibinfo{person}{Kristen Grauman}.} \bibinfo{year}{2012}\natexlab{}.
\newblock \showarticletitle{Geodesic flow kernel for unsupervised domain adaptation}. In \bibinfo{booktitle}{\emph{2012 IEEE conference on computer vision and pattern recognition}}. IEEE, \bibinfo{pages}{2066--2073}.
\newblock


\bibitem[Gopalakrishnan et~al\mbox{.}(2017)]%
        {gopalakrishnan2017deep}
\bibfield{author}{\bibinfo{person}{Kasthurirangan Gopalakrishnan}, \bibinfo{person}{Siddhartha~K Khaitan}, \bibinfo{person}{Alok Choudhary}, {and} \bibinfo{person}{Ankit Agrawal}.} \bibinfo{year}{2017}\natexlab{}.
\newblock \showarticletitle{Deep convolutional neural networks with transfer learning for computer vision-based data-driven pavement distress detection}.
\newblock \bibinfo{journal}{\emph{Construction and building materials}}  \bibinfo{volume}{157} (\bibinfo{year}{2017}), \bibinfo{pages}{322--330}.
\newblock


\bibitem[Gorelick et~al\mbox{.}(2007)]%
        {gorelick2007actions}
\bibfield{author}{\bibinfo{person}{Lena Gorelick}, \bibinfo{person}{Moshe Blank}, \bibinfo{person}{Eli Shechtman}, \bibinfo{person}{Michal Irani}, {and} \bibinfo{person}{Ronen Basri}.} \bibinfo{year}{2007}\natexlab{}.
\newblock \showarticletitle{Actions as space-time shapes}.
\newblock \bibinfo{journal}{\emph{IEEE transactions on pattern analysis and machine intelligence}} \bibinfo{volume}{29}, \bibinfo{number}{12} (\bibinfo{year}{2007}), \bibinfo{pages}{2247--2253}.
\newblock


\bibitem[Guo et~al\mbox{.}(2019)]%
        {guo2019spottune}
\bibfield{author}{\bibinfo{person}{Yunhui Guo}, \bibinfo{person}{Honghui Shi}, \bibinfo{person}{Abhishek Kumar}, \bibinfo{person}{Kristen Grauman}, \bibinfo{person}{Tajana Rosing}, {and} \bibinfo{person}{Rogerio Feris}.} \bibinfo{year}{2019}\natexlab{}.
\newblock \showarticletitle{Spottune: transfer learning through adaptive fine-tuning}. In \bibinfo{booktitle}{\emph{Proceedings of the IEEE/CVF conference on computer vision and pattern recognition}}. \bibinfo{pages}{4805--4814}.
\newblock


\bibitem[Haque(2023)]%
        {cnn-linkedin}
\bibfield{author}{\bibinfo{person}{Kh.~Nafizul Haque}.} \bibinfo{year}{2023}\natexlab{}.
\newblock \bibinfo{title}{{What is Convolutional Neural Network — CNN (Deep Learning)}}.
\newblock \bibinfo{howpublished}{\url{https://www.linkedin.com/pulse/what-convolutional-neural-network-cnn-deep-learning-nafiz-shahriar}}.
\newblock
\newblock
\shownote{Accessed: 2024-08-21}.


\bibitem[He et~al\mbox{.}(2016)]%
        {he2016deep}
\bibfield{author}{\bibinfo{person}{Kaiming He}, \bibinfo{person}{Xiangyu Zhang}, \bibinfo{person}{Shaoqing Ren}, {and} \bibinfo{person}{Jian Sun}.} \bibinfo{year}{2016}\natexlab{}.
\newblock \showarticletitle{Deep residual learning for image recognition}. In \bibinfo{booktitle}{\emph{Proceedings of the IEEE conference on computer vision and pattern recognition}}. \bibinfo{pages}{770--778}.
\newblock


\bibitem[Hridoy et~al\mbox{.}(2021)]%
        {hridoy2021computer}
\bibfield{author}{\bibinfo{person}{Rashidul~Hasan Hridoy}, \bibinfo{person}{Fatema Akter}, {and} \bibinfo{person}{Aniruddha Rakshit}.} \bibinfo{year}{2021}\natexlab{}.
\newblock \showarticletitle{Computer vision based skin disorder recognition using EfficientNet: A transfer learning approach}. In \bibinfo{booktitle}{\emph{2021 International Conference on Information Technology (ICIT)}}. IEEE, \bibinfo{pages}{482--487}.
\newblock


\bibitem[Huh et~al\mbox{.}(2016)]%
        {huh2016makes}
\bibfield{author}{\bibinfo{person}{Minyoung Huh}, \bibinfo{person}{Pulkit Agrawal}, {and} \bibinfo{person}{Alexei~A Efros}.} \bibinfo{year}{2016}\natexlab{}.
\newblock \showarticletitle{What makes ImageNet good for transfer learning?}
\newblock \bibinfo{journal}{\emph{arXiv preprint arXiv:1608.08614}} (\bibinfo{year}{2016}).
\newblock


\bibitem[IBM(2024)]%
        {ibm-cnn}
\bibfield{author}{\bibinfo{person}{IBM}.} \bibinfo{year}{2024}\natexlab{}.
\newblock \bibinfo{title}{{What are convolutional neural networks?}}
\newblock \bibinfo{howpublished}{\url{ibm.com/topics/convolutional-neural-networks}}.
\newblock
\newblock
\shownote{Accessed: 2024-07-05}.


\bibitem[Jaiswal et~al\mbox{.}(2021)]%
        {jaiswal2021classification}
\bibfield{author}{\bibinfo{person}{Aayush Jaiswal}, \bibinfo{person}{Neha Gianchandani}, \bibinfo{person}{Dilbag Singh}, \bibinfo{person}{Vijay Kumar}, {and} \bibinfo{person}{Manjit Kaur}.} \bibinfo{year}{2021}\natexlab{}.
\newblock \showarticletitle{Classification of the COVID-19 infected patients using DenseNet201 based deep transfer learning}.
\newblock \bibinfo{journal}{\emph{Journal of Biomolecular Structure and Dynamics}} \bibinfo{volume}{39}, \bibinfo{number}{15} (\bibinfo{year}{2021}), \bibinfo{pages}{5682--5689}.
\newblock


\bibitem[J{\'e}gou et~al\mbox{.}(2017)]%
        {jegou2017one}
\bibfield{author}{\bibinfo{person}{Simon J{\'e}gou}, \bibinfo{person}{Michal Drozdzal}, \bibinfo{person}{David Vazquez}, \bibinfo{person}{Adriana Romero}, {and} \bibinfo{person}{Yoshua Bengio}.} \bibinfo{year}{2017}\natexlab{}.
\newblock \showarticletitle{The one hundred layers tiramisu: Fully convolutional densenets for semantic segmentation}. In \bibinfo{booktitle}{\emph{Proceedings of the IEEE conference on computer vision and pattern recognition workshops}}. \bibinfo{pages}{11--19}.
\newblock


\bibitem[Karimi et~al\mbox{.}(2021)]%
        {karimi2021transfer}
\bibfield{author}{\bibinfo{person}{Davood Karimi}, \bibinfo{person}{Simon~K Warfield}, {and} \bibinfo{person}{Ali Gholipour}.} \bibinfo{year}{2021}\natexlab{}.
\newblock \showarticletitle{Transfer learning in medical image segmentation: New insights from analysis of the dynamics of model parameters and learned representations}.
\newblock \bibinfo{journal}{\emph{Artificial Intelligence in Medicine}}  \bibinfo{volume}{116} (\bibinfo{year}{2021}), \bibinfo{pages}{102078}.
\newblock


\bibitem[Kentsch et~al\mbox{.}(2020)]%
        {kentsch2020computer}
\bibfield{author}{\bibinfo{person}{Sarah Kentsch}, \bibinfo{person}{Maximo~Larry Lopez~Caceres}, \bibinfo{person}{Daniel Serrano}, \bibinfo{person}{Ferran Roure}, {and} \bibinfo{person}{Yago Diez}.} \bibinfo{year}{2020}\natexlab{}.
\newblock \showarticletitle{Computer vision and deep learning techniques for the analysis of drone-acquired forest images, a transfer learning study}.
\newblock \bibinfo{journal}{\emph{Remote Sensing}} \bibinfo{volume}{12}, \bibinfo{number}{8} (\bibinfo{year}{2020}), \bibinfo{pages}{1287}.
\newblock


\bibitem[Kitchenham(2004)]%
        {kitchenham2004procedures}
\bibfield{author}{\bibinfo{person}{Barbara Kitchenham}.} \bibinfo{year}{2004}\natexlab{}.
\newblock \showarticletitle{Procedures for performing systematic reviews}.
\newblock \bibinfo{journal}{\emph{Keele, UK, Keele University}} \bibinfo{volume}{33}, \bibinfo{number}{2004} (\bibinfo{year}{2004}), \bibinfo{pages}{1--26}.
\newblock


\bibitem[Kotsiantis et~al\mbox{.}(2006)]%
        {kotsiantis2006machine}
\bibfield{author}{\bibinfo{person}{Sotiris~B Kotsiantis}, \bibinfo{person}{Ioannis~D Zaharakis}, {and} \bibinfo{person}{Panayiotis~E Pintelas}.} \bibinfo{year}{2006}\natexlab{}.
\newblock \showarticletitle{Machine learning: a review of classification and combining techniques}.
\newblock \bibinfo{journal}{\emph{Artificial Intelligence Review}} \bibinfo{volume}{26}, \bibinfo{number}{3} (\bibinfo{year}{2006}), \bibinfo{pages}{159--190}.
\newblock


\bibitem[Krizhevsky et~al\mbox{.}(2017)]%
        {krizhevsky2017imagenet}
\bibfield{author}{\bibinfo{person}{Alex Krizhevsky}, \bibinfo{person}{Ilya Sutskever}, {and} \bibinfo{person}{Geoffrey~E Hinton}.} \bibinfo{year}{2017}\natexlab{}.
\newblock \showarticletitle{Imagenet classification with deep convolutional neural networks}.
\newblock \bibinfo{journal}{\emph{Commun. ACM}} \bibinfo{volume}{60}, \bibinfo{number}{6} (\bibinfo{year}{2017}), \bibinfo{pages}{84--90}.
\newblock


\bibitem[Labovitz(2019)]%
        {labovitz2019internet}
\bibfield{author}{\bibinfo{person}{Craig Labovitz}.} \bibinfo{year}{2019}\natexlab{}.
\newblock \showarticletitle{Internet Traffic 2009-2019}. In \bibinfo{booktitle}{\emph{Proc. Asia Pacific Regional Internet Conf. Operational Technologies}}.
\newblock


\bibitem[Long et~al\mbox{.}(2015)]%
        {long2015learning}
\bibfield{author}{\bibinfo{person}{Mingsheng Long}, \bibinfo{person}{Yue Cao}, \bibinfo{person}{Jianmin Wang}, {and} \bibinfo{person}{Michael Jordan}.} \bibinfo{year}{2015}\natexlab{}.
\newblock \showarticletitle{Learning transferable features with deep adaptation networks}. In \bibinfo{booktitle}{\emph{International conference on machine learning}}. PMLR, \bibinfo{pages}{97--105}.
\newblock


\bibitem[Milletari et~al\mbox{.}(2016)]%
        {milletari2016v}
\bibfield{author}{\bibinfo{person}{Fausto Milletari}, \bibinfo{person}{Nassir Navab}, {and} \bibinfo{person}{Seyed-Ahmad Ahmadi}.} \bibinfo{year}{2016}\natexlab{}.
\newblock \showarticletitle{V-net: Fully convolutional neural networks for volumetric medical image segmentation}. In \bibinfo{booktitle}{\emph{2016 fourth international conference on 3D vision (3DV)}}. Ieee, \bibinfo{pages}{565--571}.
\newblock


\bibitem[Mitra et~al\mbox{.}(2024a)]%
        {mitra2024use}
\bibfield{author}{\bibinfo{person}{Shaswata Mitra}, \bibinfo{person}{Trisha Chakraborty}, \bibinfo{person}{Subash Neupane}, \bibinfo{person}{Aritran Piplai}, {and} \bibinfo{person}{Sudip Mittal}.} \bibinfo{year}{2024}\natexlab{a}.
\newblock \showarticletitle{Use of Graph Neural Networks in Aiding Defensive Cyber Operations}.
\newblock \bibinfo{journal}{\emph{arXiv preprint arXiv:2401.05680}} (\bibinfo{year}{2024}).
\newblock


\bibitem[Mitra et~al\mbox{.}(2024b)]%
        {mitra2024localintel}
\bibfield{author}{\bibinfo{person}{Shaswata Mitra}, \bibinfo{person}{Subash Neupane}, \bibinfo{person}{Trisha Chakraborty}, \bibinfo{person}{Sudip Mittal}, \bibinfo{person}{Aritran Piplai}, \bibinfo{person}{Manas Gaur}, {and} \bibinfo{person}{Shahram Rahimi}.} \bibinfo{year}{2024}\natexlab{b}.
\newblock \showarticletitle{Localintel: Generating organizational threat intelligence from global and local cyber knowledge}.
\newblock \bibinfo{journal}{\emph{arXiv preprint arXiv:2401.10036}} (\bibinfo{year}{2024}).
\newblock


\bibitem[Mitra et~al\mbox{.}(2023)]%
        {mitra2023survey}
\bibfield{author}{\bibinfo{person}{Shaswata Mitra}, \bibinfo{person}{Stephen~A Torri}, {and} \bibinfo{person}{Sudip Mittal}.} \bibinfo{year}{2023}\natexlab{}.
\newblock \showarticletitle{Survey of malware analysis through control flow graph using machine learning}. In \bibinfo{booktitle}{\emph{2023 IEEE 22nd International Conference on Trust, Security and Privacy in Computing and Communications (TrustCom)}}. IEEE, \bibinfo{pages}{1554--1561}.
\newblock


\bibitem[Moore et~al\mbox{.}(2023)]%
        {moore2023ura}
\bibfield{author}{\bibinfo{person}{Charles Moore}, \bibinfo{person}{Shaswata Mitra}, \bibinfo{person}{Nisha Pillai}, \bibinfo{person}{Marc Moore}, \bibinfo{person}{Sudip Mittal}, \bibinfo{person}{Cindy Bethel}, {and} \bibinfo{person}{Jingdao Chen}.} \bibinfo{year}{2023}\natexlab{}.
\newblock \showarticletitle{Ura*: Uncertainty-aware path planning using image-based aerial-to-ground traversability estimation for off-road environments}.
\newblock \bibinfo{journal}{\emph{arXiv preprint arXiv:2309.08814}} (\bibinfo{year}{2023}).
\newblock


\bibitem[Munia et~al\mbox{.}(2017)]%
        {munia2017automatic}
\bibfield{author}{\bibinfo{person}{Tamanna~TK Munia}, \bibinfo{person}{Intisar~Rizwan i Haque}, \bibinfo{person}{Abby Aymond}, \bibinfo{person}{Nicholas MacKinnon}, \bibinfo{person}{Daniel~L Farkas}, \bibinfo{person}{Minhal Al-Hashim}, \bibinfo{person}{Fartash Vasefi}, {and} \bibinfo{person}{Reza Fazel-Rezai}.} \bibinfo{year}{2017}\natexlab{}.
\newblock \showarticletitle{Automatic clustering-based segmentation and plaque localization in psoriasis digital images}. In \bibinfo{booktitle}{\emph{2017 IEEE Healthcare Innovations and Point of Care Technologies (HI-POCT)}}. IEEE, \bibinfo{pages}{113--116}.
\newblock


\bibitem[Neupane et~al\mbox{.}(2024a)]%
        {neupane2024security}
\bibfield{author}{\bibinfo{person}{Subash Neupane}, \bibinfo{person}{Shaswata Mitra}, \bibinfo{person}{Ivan~A Fernandez}, \bibinfo{person}{Swayamjit Saha}, \bibinfo{person}{Sudip Mittal}, \bibinfo{person}{Jingdao Chen}, \bibinfo{person}{Nisha Pillai}, {and} \bibinfo{person}{Shahram Rahimi}.} \bibinfo{year}{2024}\natexlab{a}.
\newblock \showarticletitle{Security Considerations in AI-Robotics: A Survey of Current Methods, Challenges, and Opportunities}.
\newblock \bibinfo{journal}{\emph{IEEE Access}} (\bibinfo{year}{2024}).
\newblock


\bibitem[Neupane et~al\mbox{.}(2024b)]%
        {neupane2024medinsight}
\bibfield{author}{\bibinfo{person}{Subash Neupane}, \bibinfo{person}{Shaswata Mitra}, \bibinfo{person}{Sudip Mittal}, \bibinfo{person}{Noorbakhsh~Amiri Golilarz}, \bibinfo{person}{Shahram Rahimi}, {and} \bibinfo{person}{Amin Amirlatifi}.} \bibinfo{year}{2024}\natexlab{b}.
\newblock \showarticletitle{MedInsight: A Multi-Source Context Augmentation Framework for Generating Patient-Centric Medical Responses using Large Language Models}.
\newblock \bibinfo{journal}{\emph{arXiv preprint arXiv:2403.08607}} (\bibinfo{year}{2024}).
\newblock


\bibitem[Neyshabur et~al\mbox{.}(2020)]%
        {neyshabur2020being}
\bibfield{author}{\bibinfo{person}{Behnam Neyshabur}, \bibinfo{person}{Hanie Sedghi}, {and} \bibinfo{person}{Chiyuan Zhang}.} \bibinfo{year}{2020}\natexlab{}.
\newblock \showarticletitle{What is being transferred in transfer learning?}
\newblock \bibinfo{journal}{\emph{Advances in neural information processing systems}}  \bibinfo{volume}{33} (\bibinfo{year}{2020}), \bibinfo{pages}{512--523}.
\newblock


\bibitem[Nielsen(2015)]%
        {nielsen2015neural}
\bibfield{author}{\bibinfo{person}{Michael~A Nielsen}.} \bibinfo{year}{2015}\natexlab{}.
\newblock \bibinfo{booktitle}{\emph{Neural networks and deep learning}}. Vol.~\bibinfo{volume}{25}.
\newblock \bibinfo{publisher}{Determination press San Francisco, CA, USA}.
\newblock


\bibitem[Pan and Yang(2010)]%
        {pan2010survey}
\bibfield{author}{\bibinfo{person}{Sinno~Jialin Pan} {and} \bibinfo{person}{Qiang Yang}.} \bibinfo{year}{2010}\natexlab{}.
\newblock \showarticletitle{A survey on transfer learning}.
\newblock \bibinfo{journal}{\emph{IEEE Transactions on knowledge and data engineering}} \bibinfo{volume}{22}, \bibinfo{number}{10} (\bibinfo{year}{2010}), \bibinfo{pages}{1345--1359}.
\newblock


\bibitem[Precup and David(2019)]%
        {precup2019nature}
\bibfield{author}{\bibinfo{person}{Radu-Emil Precup} {and} \bibinfo{person}{Radu-Codrut David}.} \bibinfo{year}{2019}\natexlab{}.
\newblock \bibinfo{booktitle}{\emph{Nature-inspired optimization algorithms for fuzzy controlled servo systems}}.
\newblock \bibinfo{publisher}{Butterworth-Heinemann}.
\newblock


\bibitem[Qin et~al\mbox{.}(2018)]%
        {qin2018autofocus}
\bibfield{author}{\bibinfo{person}{Yao Qin}, \bibinfo{person}{Konstantinos Kamnitsas}, \bibinfo{person}{Siddharth Ancha}, \bibinfo{person}{Jay Nanavati}, \bibinfo{person}{Garrison Cottrell}, \bibinfo{person}{Antonio Criminisi}, {and} \bibinfo{person}{Aditya Nori}.} \bibinfo{year}{2018}\natexlab{}.
\newblock \showarticletitle{Autofocus layer for semantic segmentation}. In \bibinfo{booktitle}{\emph{International conference on medical image computing and computer-assisted intervention}}. Springer, \bibinfo{pages}{603--611}.
\newblock


\bibitem[Rathod et~al\mbox{.}(2018)]%
        {rathod2018diagnosis}
\bibfield{author}{\bibinfo{person}{Jainesh Rathod}, \bibinfo{person}{Vishal Waghmode}, \bibinfo{person}{Aniruddh Sodha}, {and} \bibinfo{person}{Praseniit Bhavathankar}.} \bibinfo{year}{2018}\natexlab{}.
\newblock \showarticletitle{Diagnosis of skin diseases using Convolutional Neural Networks}. In \bibinfo{booktitle}{\emph{2018 second international conference on electronics, communication and aerospace technology (ICECA)}}. IEEE, \bibinfo{pages}{1048--1051}.
\newblock


\bibitem[Ratul et~al\mbox{.}(2020)]%
        {ratul2020skin}
\bibfield{author}{\bibinfo{person}{Md~Aminur~Rab Ratul}, \bibinfo{person}{M~Hamed Mozaffari}, \bibinfo{person}{Won-Sook Lee}, {and} \bibinfo{person}{Enea Parimbelli}.} \bibinfo{year}{2020}\natexlab{}.
\newblock \showarticletitle{Skin lesions classification using deep learning based on dilated convolution}.
\newblock \bibinfo{journal}{\emph{BioRxiv}} (\bibinfo{year}{2020}), \bibinfo{pages}{860700}.
\newblock


\bibitem[Russell and Norvig(2016)]%
        {russell2016artificial}
\bibfield{author}{\bibinfo{person}{Stuart~J Russell} {and} \bibinfo{person}{Peter Norvig}.} \bibinfo{year}{2016}\natexlab{}.
\newblock \bibinfo{booktitle}{\emph{Artificial intelligence: a modern approach}}.
\newblock \bibinfo{publisher}{Pearson}.
\newblock


\bibitem[Shanthi et~al\mbox{.}(2020)]%
        {shanthi2020automatic}
\bibfield{author}{\bibinfo{person}{T Shanthi}, \bibinfo{person}{RS Sabeenian}, {and} \bibinfo{person}{R Anand}.} \bibinfo{year}{2020}\natexlab{}.
\newblock \showarticletitle{Automatic diagnosis of skin diseases using convolution neural network}.
\newblock \bibinfo{journal}{\emph{Microprocessors and Microsystems}}  \bibinfo{volume}{76} (\bibinfo{year}{2020}), \bibinfo{pages}{103074}.
\newblock


\bibitem[Szeliski(2022)]%
        {szeliski2022computer}
\bibfield{author}{\bibinfo{person}{Richard Szeliski}.} \bibinfo{year}{2022}\natexlab{}.
\newblock \bibinfo{booktitle}{\emph{Computer vision: algorithms and applications}}.
\newblock \bibinfo{publisher}{Springer Nature}.
\newblock


\bibitem[Tan and Le(2019)]%
        {tan2019efficientnet}
\bibfield{author}{\bibinfo{person}{Mingxing Tan} {and} \bibinfo{person}{Quoc Le}.} \bibinfo{year}{2019}\natexlab{}.
\newblock \showarticletitle{Efficientnet: Rethinking model scaling for convolutional neural networks}. In \bibinfo{booktitle}{\emph{International conference on machine learning}}. PMLR, \bibinfo{pages}{6105--6114}.
\newblock


\bibitem[Voulodimos et~al\mbox{.}(2018)]%
        {voulodimos2018deep}
\bibfield{author}{\bibinfo{person}{Athanasios Voulodimos}, \bibinfo{person}{Nikolaos Doulamis}, \bibinfo{person}{Anastasios Doulamis}, {and} \bibinfo{person}{Eftychios Protopapadakis}.} \bibinfo{year}{2018}\natexlab{}.
\newblock \showarticletitle{Deep learning for computer vision: A brief review}.
\newblock \bibinfo{journal}{\emph{Computational intelligence and neuroscience}}  \bibinfo{volume}{2018} (\bibinfo{year}{2018}).
\newblock


\bibitem[Wang et~al\mbox{.}(2020)]%
        {wang2020deep}
\bibfield{author}{\bibinfo{person}{Jingdong Wang}, \bibinfo{person}{Ke Sun}, \bibinfo{person}{Tianheng Cheng}, \bibinfo{person}{Borui Jiang}, \bibinfo{person}{Chaorui Deng}, \bibinfo{person}{Yang Zhao}, \bibinfo{person}{Dong Liu}, \bibinfo{person}{Yadong Mu}, \bibinfo{person}{Mingkui Tan}, \bibinfo{person}{Xinggang Wang}, {et~al\mbox{.}}} \bibinfo{year}{2020}\natexlab{}.
\newblock \showarticletitle{Deep high-resolution representation learning for visual recognition}.
\newblock \bibinfo{journal}{\emph{IEEE transactions on pattern analysis and machine intelligence}} \bibinfo{volume}{43}, \bibinfo{number}{10} (\bibinfo{year}{2020}), \bibinfo{pages}{3349--3364}.
\newblock


\bibitem[Weiss et~al\mbox{.}(2016)]%
        {weiss2016survey}
\bibfield{author}{\bibinfo{person}{Karl Weiss}, \bibinfo{person}{Taghi~M Khoshgoftaar}, {and} \bibinfo{person}{DingDing Wang}.} \bibinfo{year}{2016}\natexlab{}.
\newblock \showarticletitle{A survey of transfer learning}.
\newblock \bibinfo{journal}{\emph{Journal of Big data}} \bibinfo{volume}{3}, \bibinfo{number}{1} (\bibinfo{year}{2016}), \bibinfo{pages}{1--40}.
\newblock


\bibitem[Zhang and Saligrama(2015)]%
        {zhang2015zero}
\bibfield{author}{\bibinfo{person}{Ziming Zhang} {and} \bibinfo{person}{Venkatesh Saligrama}.} \bibinfo{year}{2015}\natexlab{}.
\newblock \showarticletitle{Zero-shot learning via semantic similarity embedding}. In \bibinfo{booktitle}{\emph{Proceedings of the IEEE international conference on computer vision}}. \bibinfo{pages}{4166--4174}.
\newblock


\bibitem[Zhou et~al\mbox{.}(2018)]%
        {zhou2018unet++}
\bibfield{author}{\bibinfo{person}{Zongwei Zhou}, \bibinfo{person}{Md~Mahfuzur Rahman~Siddiquee}, \bibinfo{person}{Nima Tajbakhsh}, {and} \bibinfo{person}{Jianming Liang}.} \bibinfo{year}{2018}\natexlab{}.
\newblock \showarticletitle{Unet++: A nested u-net architecture for medical image segmentation}.
\newblock In \bibinfo{booktitle}{\emph{Deep learning in medical image analysis and multimodal learning for clinical decision support}}. \bibinfo{publisher}{Springer}, \bibinfo{pages}{3--11}.
\newblock


\bibitem[Zhuang et~al\mbox{.}(2020)]%
        {zhuang2020comprehensive}
\bibfield{author}{\bibinfo{person}{Fuzhen Zhuang}, \bibinfo{person}{Zhiyuan Qi}, \bibinfo{person}{Keyu Duan}, \bibinfo{person}{Dongbo Xi}, \bibinfo{person}{Yongchun Zhu}, \bibinfo{person}{Hengshu Zhu}, \bibinfo{person}{Hui Xiong}, {and} \bibinfo{person}{Qing He}.} \bibinfo{year}{2020}\natexlab{}.
\newblock \showarticletitle{A comprehensive survey on transfer learning}.
\newblock \bibinfo{journal}{\emph{Proc. IEEE}} \bibinfo{volume}{109}, \bibinfo{number}{1} (\bibinfo{year}{2020}), \bibinfo{pages}{43--76}.
\newblock


\end{thebibliography}

\end{document}